\pdfoutput=1

\documentclass[11pt]{article}

\usepackage[final]{acl}

\usepackage{times}
\usepackage{latexsym}
\usepackage{hyperref}

\usepackage[T1]{fontenc}
\usepackage{arabtex}
\usepackage{utf8} 
\usepackage[utf8]{inputenc} 

\usepackage{microtype}

\usepackage{inconsolata}

\usepackage{graphicx}
\usepackage{times}
\usepackage{tabu}
\usepackage{latexsym}
\usepackage{stmaryrd}
\usepackage{amsmath}
\usepackage{multirow}
\usepackage{bbm}
\usepackage{graphicx}
\usepackage{amssymb}
\usepackage{booktabs}
\usepackage{algpseudocode}
\usepackage{algorithm}
\usepackage{graphicx}
\usepackage{caption}
\usepackage{subcaption}
\usepackage{tabularx}
\usepackage{colortbl}
\usepackage{xcolor}
\usepackage{tablefootnote}
\usepackage{enumitem}  
\usepackage{lipsum}
\usepackage{xspace}
\usepackage{bbding}
\usepackage{pifont}
\usepackage{arydshln}
\usepackage{array}
\usepackage{cleveref}
\newcommand{\PreserveBackslash}[1]{\let\temp=\\#1\let\\=\temp}
\newcolumntype{C}[1]{>{\PreserveBackslash\centering}p{#1}}
\newcolumntype{R}[1]{>{\PreserveBackslash\raggedleft}p{#1}}
\newcolumntype{L}[1]{>{\PreserveBackslash\raggedright}p{#1}}
\newcommand{\datasetname}{\texttt{ArabCulture}\xspace}

\definecolor{mygreen}{RGB}{217, 234, 211}
\definecolor{myred}{RGB}{244, 204, 204}

\newcommand{\ok}{\cellcolor{mygreen}}
\newcommand{\no}{\cellcolor{myred}}

\title{Commonsense Reasoning in Arab Culture}

\author{Abdelrahman Sadallah$^{1}$ \quad Junior Cedric Tonga$^{1}$ \quad Khalid Almubarak$^{2,5}$ \\  \textbf{Saeed Almheiri}$^{1}$ \quad  \textbf{Farah Atif}$^{1}$ \,   \textbf{Chatrine Qwaider}$^{1}$ \quad \textbf{Karima Kadaoui}$^{1}$ \\   \textbf{Sara Shatnawi}$^{3}$  \quad  \textbf{Yaser Alesh}$^{4}$  \quad \textbf{Fajri Koto}$^{1}$ \\ 
$^{1}$Mohamed bin Zayed University of Artificial Intelligence \\
        $^{2}$SDAIA \, $^{3}$Al-Balqa Applied University \\  $^{4}$Khalifa University \, $^{5}$HUMAIN, Data and AI Models\\ 
	\texttt{\small \{abdelrahman.sadallah,fajri.koto\}@mbzuai.ac.ae 
	} 
}

\begin{document}
\setcode{utf8}

\maketitle

\begin{abstract}
Despite progress in Arabic large language models, such as Jais and AceGPT, their evaluation on commonsense reasoning has largely relied on machine-translated datasets, which lack cultural depth and may introduce Anglocentric biases. Commonsense reasoning is shaped by geographical and cultural contexts, and existing English datasets fail to capture the diversity of the Arab world. To address this, we introduce \datasetname, a commonsense reasoning dataset in Modern Standard Arabic (MSA), covering cultures of 13 countries across the Gulf, Levant, North Africa, and the Nile Valley. The dataset was built from scratch by engaging native speakers to write and validate culturally relevant questions for their respective countries. \datasetname spans 12 daily life domains with 54 fine-grained subtopics, reflecting various aspects of social norms, traditions, and everyday experiences. Zero-shot evaluations show that open-weight language models with up to 32B parameters struggle to comprehend diverse Arab cultures, with performance varying across regions. These findings highlight the need for more culturally aware models and datasets tailored to the Arabic-speaking world.\footnote{\datasetname{} can be accessed at \url{https://huggingface.co/datasets/MBZUAI/ArabCulture}}
\end{list}
\end{abstract}


\section{Introduction}
\label{sec:intro}
Commonsense reasoning is the ability to make judgments and inferences based on everyday human knowledge and experiences \cite{sap-etal-2020-commonsense}. It is a fundamental aspect of human cognition and has been extensively studied in the context of large language models (LLMs) \citep{openai2024gpt4technicalreport, grattafiori2024llama3herdmodels, Liu2023LLM360TF}. However, commonsense reasoning is not universal—it is shaped by culture, which encompasses the shared knowledge, values, customs, and behaviors that define a society \citep{macionis2012sociology,giddens2014essential}.
\begin{figure}[H]
    \centering
    \includegraphics[width=\linewidth]{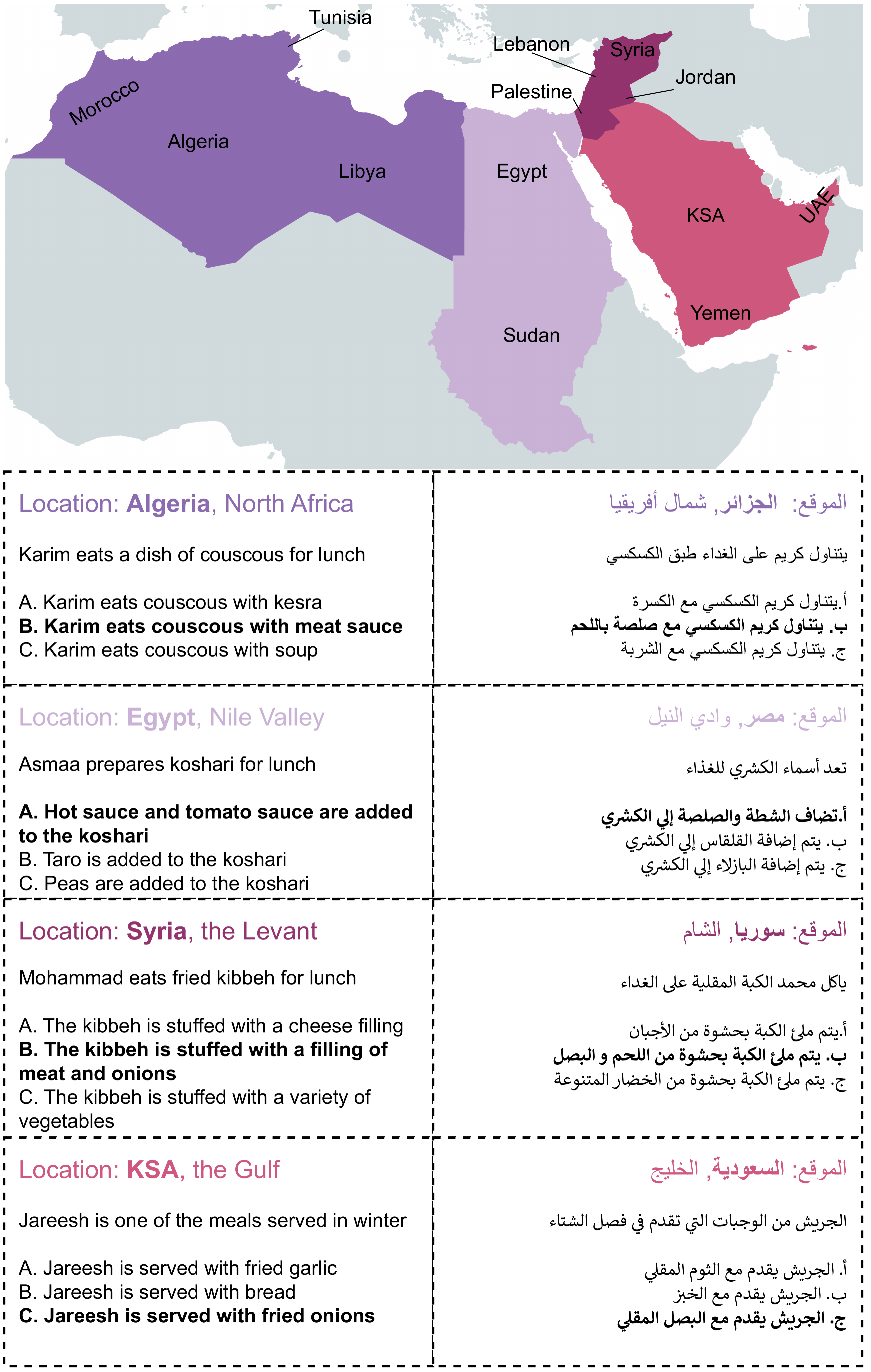} 
    \caption{\datasetname covers four regions across the Middle East and North Africa, spanning 13 countries. The highlighted areas on the map represent the regions included in \datasetname. We present example questions for the \textit{Lunch} category from Algeria, Egypt, Syria, and KSA, each with three answer choices, with the correct answer marked in bold. English translations are provided for illustration.}
    \label{fig:region_examples}
\end{figure}

The Arab world, home to approximately 456 million people \citep{diab-etal-2017-nlp,shoufan-alameri-2015-natural}, is characterized by its linguistic unity through Modern Standard Arabic (MSA) while encompassing diverse traditions, religions, and customs \citep{arabworld}. This cultural diversity influences not only social interactions but also reasoning patterns, making it crucial to develop evaluation benchmarks that reflect these variations. However, most existing commonsense reasoning datasets are developed with Western-centric assumptions, limiting their applicability to Arabic-speaking societies.

Despite recent advancements in Arabic LLMs \citep{sengupta2023jais, huang2024acegpt,silma_01_2024,bari2024allam}, their evaluation has largely relied on machine-translated datasets originally created in English. Many commonsense reasoning benchmarks \citep{Bisk2019PIQARA, socialiqa} fail to capture Arab cultural perspectives, as simple translations do not account for region-specific knowledge, potentially introducing bias. Given the significant cultural variations across the Arab world, these datasets do not provide a holistic measure of Arabic LLMs' ability to reason within culturally specific contexts. This raises an important question: \textit{To what extent can existing LLMs accurately reason about commonsense knowledge in diverse cultural settings, particularly in the Arab world?}

To address this gap, we introduce \datasetname, a commonsense reasoning dataset specifically designed to assess the cultural knowledge of Arabic LLMs. The dataset consists of 3,482 questions written in Modern Standard Arabic (MSA), covering 13 countries across the Gulf, Levant, North Africa, and the Nile Valley (see Figure~\ref{fig:region_examples}). It spans 12 major domains and 54 fine-grained subtopics, reflecting various aspects of social norms, traditions, and daily life in the Arab world. Unlike existing benchmarks, which often rely on translated datasets, \datasetname was built from scratch by directly engaging native speakers to write culturally relevant questions for their respective countries. We carefully implemented quality control measures throughout the dataset creation process, including rigorous validation steps to maintain accuracy, relevance, and cultural sensitivity.

We evaluate a range of closed-weight and open-weight Arabic and multilingual LLMs in a zero-shot setting to assess their cultural commonsense reasoning capabilities. Inspired by~\citet{koto-etal-2024-indoculture}, we frame the task in two ways: multiple-choice questions (MCQ) and completion tasks. Additionally, we introduce three levels of location-based contextual grounding: (1) no additional location information, (2) specifying the broader region (e.g., Gulf or Levant), and (3) specifying the exact country along with its regional classification. This setup allows us to analyze how effectively LLMs incorporate geographical and cultural cues in their reasoning.

Our results show that even LLMs with up to 32B parameters struggle with cultural commonsense reasoning, with performance varying significantly across regions. We conduct a detailed analysis of the best-performing models, identifying strengths and weaknesses across different cultural contexts. We also conducted a small manual experiment to test the models' ability to explain their chosen answers. Additionally, we explore whether enriching prompts with cultural facts improves performance in smaller language models, finding that while it helps in some cases, it does not provide a universal solution.

\begin{table*}[t]
    \centering
    \renewcommand{\arraystretch}{1.3} 
    \setlength{\tabcolsep}{4pt} 
    \resizebox{1\textwidth}{!}{ \begin{tabular}{lrlcccccc}
            \toprule
            \textbf{Dataset} & \textbf{Size} & \textbf{Data Construction Method} & \textbf{Cultural?} & \textbf{Location?} & \textbf{\#Topic} & \textbf{\#Country} & \textbf{Reasoning?} \\
            \midrule
            \datasetname{} (\textbf{Ours}) & 3,482 & Manually built,  validated by native & \checkmark & \checkmark &  54 & 13 & \checkmark \\
            AraDiCE-Culture \cite{mousi-etal-2025-aradice} & 180 & Manually built, validated by native & \checkmark & \checkmark & 9 &  1 & -- \\
            AraDiCE-WinoGrande \cite{mousi-etal-2025-aradice} & 1,267 & Machine-translated, post-edited & -- & -- & -- & --& \checkmark \\
            AraDiCE-PIQA \cite{mousi-etal-2025-aradice} & 1,838 & Machine-translated, post-edited & -- & -- & -- & --& \checkmark \\
            AraDiCE-OpenBookQA \cite{mousi-etal-2025-aradice} & 500 & Machine-translated, post-edited &  -- & -- & -- & -- & \checkmark \\
            AlGhafa (COPA Ar) \cite{almazrouei-etal-2023-alghafa} & 89 & Machine-translated, verified by humans & -- & -- & -- & -- & \checkmark \\
            ACVA \cite{huang2024acegpt} & 2,486 & ChatGPT generated, verified by humans & \checkmark &  -- & 50 & -- & -- \\
            \bottomrule
        \end{tabular}
    }
    \caption{Comparison of our dataset with other Arabic cultural commonsense reasoning datasets. The metadata includes\textbf{ Size} (number of Arabic instances), \textbf{Cultural?} (whether the data considers cultural nuances), \textbf{Location?} ( whether the data includes fine-grained location information, such as regions and countries per region), \textbf{\#Topic} (number of fine-grained topics covered), \textbf{\#Country} (total number of countries across all regions) and \textbf{Reasoning?} (whether the data emphasizes commonsense reasoning or not). }
    \label{tab:dataset_comparison} 
\end{table*}

\section{Related Work}
\subsection{Commonsense Reasoning in English} 


Early research on commonsense reasoning primarily focused on linguistic reasoning, as seen in the Winograd Schema Challenge~\citep{levesque2012winograd} and Winogrande~\citep{sakaguchi2019adversarial}, which evaluate pronoun coreference resolution within social and linguistic contexts. Other works have explored physical commonsense reasoning, assessing model's understanding of real-world properties and relationships~\citep{Bisk2019PIQARA}, as well as social reasoning, where models are tested on their ability to interpret human emotions, actions, and social norms \citep{sap2019socialiqa}. Research has also expanded into numerical \cite{lin-etal-2020-birds,akhtar-etal-2023-exploring}, temporal \cite{tan-etal-2023-towards}, and causal reasoning \cite{roemmele2011choice,du-etal-2022-e}, broadening the scope of commonsense evaluation. However, these benchmarks are all developed in English and shaped by Western cultural assumptions, limiting their applicability to the Arab world.

\subsection{Arabic Large Language Models and Their Evaluation on Commonsense Reasoning}
A limited number of Arabic language models have been developed with more than 7B parameters, all of which are decoder-only architectures. These include \texttt{JAIS}~\citep{sengupta2023jais}, \texttt{Fanar}~\citep{fanar}, \texttt{AceGPT}~\citep{huang2024acegpt}, and \texttt{ALLAM}~\citep{bari2024allam}. Their evaluation of commonsense reasoning has been primarily based on machine-translated datasets from English to Arabic (e.g., ~\citet{Tawalbeh2020IsTS, al2021commonsense}). Although this approach provides a useful reference, it does not offer a comprehensive assessment of how well these models capture culturally grounded commonsense knowledge.

Much of the recent Arabic-centric benchmarks have focused on classic NLP tasks, including syntax, semantics, and question answering, as seen in \texttt{LaraBench}~\citep{abdelali2024larabench}, natural language generation in \texttt{DOLPHIN}~\citep{elmadany2023dolphin}, and natural language understanding in \texttt{ORCA}~\citep{elmadany-etal-2023-orca}. Only a few studies have shifted their focus toward knowledge-intensive tasks and reasoning abilities. Among them, \texttt{ArabicMMLU}~\citep{koto-etal-2024-arabicmmlu} compiles exam questions from different education levels across Arabic-speaking countries, offering a broad knowledge assessment but placing less emphasis on cultural reasoning. 

Table~\ref{tab:dataset_comparison} compares \datasetname{} with related Arabic datasets. Most existing datasets are derived from machine translation with post-editing, prioritizing linguistic accuracy over cultural relevance. While some assess reasoning, they often lack cultural grounding, location metadata, and fine-grained topic categorization. ACVA~\citep{huang2024acegpt}, generated using ChatGPT~\citep{ouyang2022training}, is not designed for reasoning evaluation. Similarly, \texttt{AraDice-Culture}~\citep{mousi-etal-2025-aradice} consists of only 180 samples and focuses on open-ended cultural knowledge rather than structured reasoning tasks. These gaps highlight the need for larger, more diverse benchmarks that better capture Arabic cultural contexts and reasoning abilities.

\section{ArabCulture Dataset}

\datasetname{} is a sentence completion task in MSA, comprising 3,482 unique instances. Each question consists of a one-sentence premise and three answer choices that are both logically and syntactically valid. As illustrated in Figure~\ref{fig:region_examples}, instances are drawn from various Arab regions.

Solving these questions requires cultural knowledge specific to the country referenced, as the correct answer aligns with culturally relevant context. This makes \datasetname{} a valuable benchmark for assessing an LLM’s ability to incorporate cultural understanding and knowledge in Arabic-language tasks.

\subsection{Dataset Construction}
\datasetname{} is built from scratch without relying on web-scraped text, minimizing the risk of training data leakage when evaluating LLMs. It is manually created and validated by native speakers from 13 Arabic-speaking countries. To further ensure quality, the authors conduct rigorous manual checks for lexical accuracy, semantic coherence, and contextual relevance.\footnote{The authors of this paper represent most of the studied countries, contributing diverse regional perspectives to the dataset.}

\paragraph{Worker requirements}
We hired 26 expert workers from 13 Arab countries, with two workers per country, that fit defined eligibility criteria: (1) The worker must be a native Arabic speaker; (2) They must have lived in the country for at least 10 years; (3) They must possess a strong understanding of local culture and traditions; (4) Their parents must also be from the country and reside there; (5) They must have at least a high school diploma, while higher degrees were considered an advantage. Among the 26 workers, 14 hold a Bachelor's degree, including seven with a Master's degree, two with a PhD, and three with a high school diploma.

All workers were required to attend a one-hour online workshop or watch a recorded session. The workshop introduced the project concept, explained task guidelines, and addressed any potential questions. To ensure a clear understanding of the task, we conducted a pilot study before the main annotation phase.

Each worker was assigned two tasks: (1) Writing instances and (2) Reviewing and verifying the work of their peer from the same country. The payment was determined based on the minimum monthly salary for data entry jobs in each worker's country, and each worker was compensated for the equivalent of four full-time working days.


\paragraph{Country Selection}
We selected countries that ensured broad geographic coverage of the Arab world while remaining within budget constraints. Priority was given to countries with larger populations and land areas, resulting in a selection of 13 countries across four regions, representing approximately 82\% of the total Arab world population. These include: (1) The Gulf: Saudi Arabia, Yemen, UAE; (2) The Levant: Syria, Jordan, Palestine, Lebanon; (3) North Africa: Morocco, Algeria, Tunisia, Libya; and (4) The Nile Valley: Egypt and Sudan.

\paragraph{Topic taxonomy}
We define 12 daily life topics with 54 fine-grained subtopics to build \datasetname{}. The topic selection is based on \citet{koto-etal-2024-indoculture} and adapted to reflect Arab regional culture. Native Arabic speakers from nine countries contributed to determining these topics, ensuring cultural relevance, diversity, and regional representation (e.g., \textit{Ramadan} traditions). Additionally, we carefully balanced the dataset by assigning an appropriate number of examples to each topic. Table \ref{tab:topics} in the Appendix provides an overview of the topics and their subtopics, which include food, weddings, holiday activities, daily activities, habits, traditional games, death, art, parenting, agriculture, family relationships, and idioms.


\paragraph{Instance Writing}
In the first stage, each worker was tasked with writing short two-sentence stories. For each entry, they were provided with a pre-defined topic and instructed to write a one-sentence premise followed by three candidate completions for the second sentence. The completions had to adhere to the following rules: (1) all had to be valid syntactic continuations of the premise, (2) none could introduce logical contradictions (e.g., ensuring consistency in topic and narrative), and (3) only one of the three sentences should be culturally accurate for the specified country. This design ensures that model predictions are influenced by cultural knowledge rather than grammatical or logical inconsistencies. Each worker was required to write 150 instances with two workers assigned per country.


\paragraph{Two-stage of Quality Control}
\label{quality-control}

In stage 1, each of the 13 countries had a designated representative involved in dataset development, all of whom are also authors of this project. After workers completed their assigned instances, the respective country representatives manually reviewed their submissions to ensure adherence to the guidelines. Linguistic errors were corrected through manual editing, but if an instance did not meet the guidelines, workers were required to revise and resubmit it.

To further ensure quality, stage 2 involved a peer validation process, where each worker reviewed their colleague’s work. The data was reformatted into multiple-choice questions, with the second sentence of each instance shuffled among three options. Workers were then asked to select the correct culturally appropriate completion and were allowed to consult external sources if unsure. If the worker selected the correct answer, it indicated agreement between annotators on the cultural validity of the instance. However, if the worker selected the wrong answer, the example was discarded, as it suggested ambiguity or cultural disagreement in the instance.




\paragraph{Country-Specific Annotation}
\label{country-specific}
Beyond quality control, we also tasked the quality control workers with annotating whether the cultural context described in an instance could be relevant to other countries. The goal was to distinguish instances that are truly unique to the designated country from those that are shared across multiple countries. To ensure accuracy, the authors of this paper conducted a second round of annotation. If an instance was marked as culturally relevant to more than one country by at least one annotator, we flagged it as Not Country-Specific ($\neg$CS); otherwise, it was labeled as Country-Specific (CS). This categorization was used in our analysis experiments to better understand the distribution of culturally unique and widely shared knowledge.


\subsection{Data Statistics}
During the instance writing phase, we initially aimed to collect 3,900 samples (26 workers $\times$ 150 samples each). However, the first quality-check round (\S \ref{quality-control}) resulted in 3,606 samples. Following the second quality control, we discarded 124 samples, leaving a final dataset of 3,482 instances. 

\begin{figure}[t]
    \centering
    \includegraphics[width=0.48\textwidth]{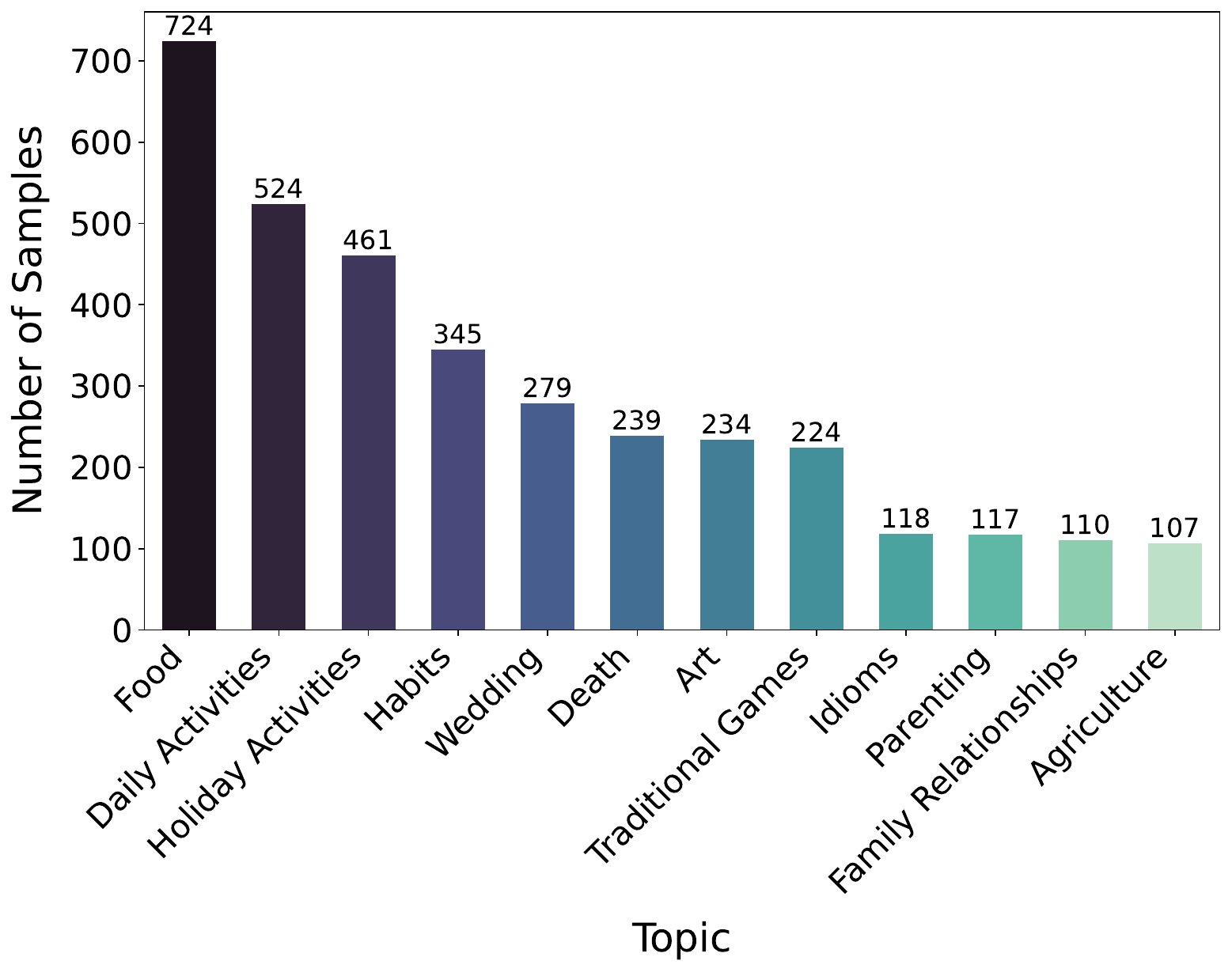}
    \caption{Total number of samples for each topic.}
    \label{fig:samples_per_topic}
    \vspace{-0.5cm}
\end{figure}

Table~\ref{tab:region-data} shows the distribution of \datasetname{} across regions and their respective countries. The overall proportion of country-specific instances is 46\%, indicating notable cultural similarities among Arab countries. In terms of word and character count, the dataset shows consistent length across countries, with an average of 32.5 words and 181 characters per instance. However, Libyan examples tend to be longer, averaging 226 characters per instance. Furthermore, Figure~\ref{fig:samples_per_topic} illustrates the total number of samples for each topic. Overall, the dataset covers a wide range of topics, with food, daily activities, and holiday activities being the most frequent, while parenting, family relationships, and agriculture are the least frequent.


\begin{center}

\begin{table}[t]
\centering
\resizebox{\columnwidth}{!}{%
\begin{tabular}{lrrrr}
\toprule
\textbf{Region} & \textbf{\#data} & \textbf{CS (\%)} & \textbf{$\mu$(words)} & \textbf{$\mu$(chars)} \\ 
\midrule
\textbf{Gulf}            & 817                 & 49.1                     & 34.6                & 188                 \\ 
KSA             & 261                 & 36.4                     & 33.2                & 185                 \\ 
UAE             & 283                 & 35.3                     & 38.1                & 205               \\ 
Yemen           & 273                 & 75.5                     & 32.4                & 172               \\ \hline
\textbf{Levant}          & 1,097                & 16.9                     & 30.2                & 170               \\
Lebanon         & 255                 & 38.8                     & 29.5                & 167                 \\
Syria           & 279                 & 16.5                     & 27.7                & 143              \\
Palestine       & 273                 & 8.4                      & 31.7                & 177               \\ 
Jordan          & 290                 & 5.9                     & 31.6                & 190               \\ \hline
\textbf{North Africa}    & 1,047                & 32.4                     & 32.4                & 179               \\ 
Tunisia         & 261                 & 31.8                     & 28.1                & 154               \\ 
Algeria         & 271                 & 27.3                     & 32.5                & 180                 \\ 
Morocco         & 276                 & 37.3                     & 28.6                & 161               \\ 
Libya           & 239                 & 33.1                     & 41.4                & 226               \\ \hline
\textbf{Nile Valley}     & 521                 & 65.5                     & 34.3                & 195               \\ 
Egypt           & 265                 & 74.3                     & 32.4                & 178               \\ 
Sudan           & 256                 & 56.2                     & 36.3                & 213               \\   \hline

\textbf{All} & 3,482 & 46.0 & 32.5 & 181 \\ \bottomrule 
\end{tabular}%
}
\caption{Overall statistics of \datasetname. CS samples represent the percentage of country-specific instances for each location. The last two columns include the average number of words and characters.   }
\label{tab:region-data}
\end{table}

\end{center}

\begin{table*}[t]
\centering
\resizebox{0.825\textwidth}{!}{%
\begin{tabular}{lcccccc}
\hline
\multirow{2}{*}{\textbf{Model (\#parameter)}} &

  \multicolumn{3}{c}{\cellcolor{blue!7}\textbf{Completion}} &
  \multicolumn{3}{c}{\cellcolor{red!7}\textbf{MCQ}} \\ \cline{2-7} 
                             & \cellcolor{blue!7}$\ell =$ None    & \cellcolor{blue!7}$\ell =$ R  & \cellcolor{blue!7}$\ell =$ R + C & \cellcolor{red!7}$\ell =$ None    & \cellcolor{red!7}$\ell =$ R  & \cellcolor{red!7}$\ell =$ R + C \\ \hline
Human                        & $-$           & $-$           & 100.0         & $-$           & $-$           & 100.0         \\
Random                                        & 33.3          & 33.3            & 33.3                      & 33.3          & 33.3            & 33.3                       \\ 
\hline

BLOOMZ (7B)                             & 31.6         & 31.4           & 31.7                     & 57.9         & 57.8           & 58.5                       \\ \hdashline
mT0$_\text{xxl}$ (14B)                                     & 27.3          & 27.6            & 27.4                      & 65.9          & 66.3            & 67.0                         \\ \hdashline
Llama-3.1 (8B)                                & 29.7         & 29.7            & 29.6                      & 35.0          & 34.7            & 34.9                       \\
Llama-3.1 Instruct (8B)                       & 30.9         & 31.0             & 31.2                      & 47.5          & 47.0            & 49.1                       \\
Llama-3 Instruct (70B)                        & 39.1         & 39.5            & 39.4                      & 34.3          & 34.3            & 34.3                       \\
Llama-3.3 Instruct (70B)                      & 39.9         & \textbf{40.6}            & \textbf{41.1}                      & 75.4          & 74.0            & 71.2                       \\ 
\hdashline
Aya-Expanse (8B)                              & 33.7         & 37.2           & 38.2                      & 39.6         & 40.7            & 41.8                      \\
Aya-Expanse (32B)                             & 37.9         & 37.9            & 39.5                      & 52.6          & 49.5            & 49.5                       \\ 
\hdashline
Gemma-2 (9B)                                  & 31.8         & 31.7           & 31.8                     & 35.2         & 34.5           & 34.5                      \\
Gemma-2 Instruct (9B)                         & 33.5         & 33.8            & 33.9                     & 58.7          & 55.3           & 57.0                      \\
Gemma-2 (27B)                                 & 32.6         & 33.0            & 33.2                      & 34.3         & 34.3           & 34.3                      \\
Gemma-2 Instruct (27B)                        & 38.0         & 38.9           & 39.8                      & 61.6         & 64.7           & 64.2                      \\ 
\hdashline
Qwen2.5 (7B)                                  & 29.0         & 31.5           & 31.8                     & 52.1         & 48.1           & 49.0                      \\
Qwen2.5 Instruct (7B)                         & 33.2         & 33.5           & 33.6                     & 53.1         & 47.9           & 48.8                      \\
Qwen2.5 (14B)                                 & 33.6         & 34.5            & 35.4                      & 55.2          & 62.5            & 61.6                       \\
Qwen2.5 Instruct (14B)                        & 36.5         & 35.9            & 37.0                      & 67.7         & 67.9            & 69.3                       \\
Qwen2.5 (32B)                                 & 34.9         & 35.6           & 35.9                     & 51.6         & 56.6           & 53.3                      \\
Qwen2.5 Instruct (32B)                        & 37.6         & 37.3           & 38.6                     & 75.2         & 75.8           & 76.5                      \\
Qwen2.5 (72B)                                 & 35.5         & 36.7           & 37.4                     & 56.1         & 51.6           & 51.8                      \\
Qwen2.5 Instruct (72B)                        & \textbf{40.1}         & 40.2           & 40.3                     & \textbf{80.1}         & \textbf{79.8}           & \textbf{80.0}                      \\ 
\hline

DeepSeek-R1-Distill-Llama (70B)  & 37.4 & 37.6 & 38.4 & 34.3 & 35.2 & 34.5 \\
DeepSeek-R1-Distill-Qwen (32B) & 34.8 & 34.7 & 35.0 & 34.3 & 34.3 & 34.3 \\
QwQ (32B)& 34.4 & 35.7 & 36.4 & 36.7 & 35.1 & 35.6 \\

\hline
Jais (13B)                                    & 39.3         & 39.0           & 39.3                      & 34.1         & 34.9            & 34.8                       \\
Jais Chat (13B)                               & 40.8         & 40.8           & 41.9                      & 58.3         & 54.1            & 54.4                       \\
Jais-v3 (30B)                                 & 39.4         & 38.4           & 39.1                      & 34.3         & 34.3            & 34.3                       \\
Jais-v3 Chat (30B)                            & 33.3         & 33.6            & 33.4                      & 60.1         & 56.2            & 54.0                       \\ 
\hdashline
SILMA Instruct (9B)                           & 32.7         & 33.0           & 33.2                      & 71.5         & 71.0            & 72.0                         \\ 
\hdashline
AceGPT-v2 (8B)                                & 32.4         & 34.0           & 34.6                     & 35.1         & 35.0           & 35.1                      \\
AceGPT-v2 Chat (8B)                           & 36.0         & 36.4           & 37.3                     & 43.1         & 39.7           & 39.3                      \\
AceGPT-v2 Chat (32B)                          & 38.5         & 39.2            & 40.0                     & \textbf{79.7}         & \textbf{79.1}           & \textbf{79.6}                      \\ 

AceGPT-v2 Chat (70B)                          & \textbf{43.2}         & \textbf{44.3}           & \textbf{44.5}                     & 61.9         & 61.7           & 62.4                      \\
\hdashline
ALLaM-Instruct-preview (7B)                           & 37.7         & 38.4           & 39.2                      & 67.4         & 72.0            & 72.6                         \\ 

\hline
GPT-4o                                        &     --          &                -- & --                          & \textbf{88.5}         & \textbf{89.6}            & \textbf{90.0}                        \\
\hline
\end{tabular}%
}
\caption{Zero-shot accuracy results for the English prompt across various models and settings. ''MCQ`` refers to the multiple-choice question evaluation method, and  $\ell$ represents the inclusion of location context (''R`` indicates the region, and ''C`` denotes the corresponding country). Bolded numbers highlight the highest score within each model group}
\label{tab:model_scores}
\end{table*}

\section{Experiment}

\subsection{Experimental Setup}

We conducted zero-shot experiments across 31 models, categorized into the following groups: (1)  20 multilingual models of various sizes, such as BLOOMZ \cite{muennighoff-etal-2023-crosslingual}, mT0-xxl \cite{muennighoff-etal-2023-crosslingual}, Llama–3 \cite{grattafiori2024llama3herdmodels}, Aya-Expanse \cite{dang2024ayaexpansecombiningresearch}, Gemma-2 \cite{Riviere2024Gemma2I}, and Qwen2.5 \cite{qwen2025qwen25technicalreport}; (2) 10 Arabic-centric models of different sizes, including Jais \cite{sengupta2023jais}, SILMA \cite{silma_01_2024}, AceGPT-v2 \cite{huang2024acegpt}, and ALLaM \cite{bari2024allamlargelanguagemodels};
(3) 3 reasoning models that are distilled from the DeepSeek-R1 model~\citep{deepseekai2025deepseekr1incentivizingreasoningcapability}. 
(4) 1 closed-weight model, GPT-4o \cite{openai2024gpt4ocard}. All experiments are done using zero temperature (greedy sampling) to enforce the models to produce factual outputs. 

We conducted experiments using both Arabic and English prompts ( Figure~\ref{fig:prompts}) and evaluated language models using two strategies: (1) sentence completion and (2) MCQ. In the sentence completion approach, we concatenate the premise with each candidate's second sentence and select the one with the highest likelihood. For MCQ, we assign alphabetical labels to the answer choices (A, B, C for English, and \<أ>\,,  \<ب>,  \<ج> for Arabic), and the selected answer corresponds to the option with the highest probability. We constructed these experiments using the LM-Evaluation-Harness Framework~\cite{eval-harness}.
Note that for the closed-weight model, we only perform MCQ-style evaluation, instructing the model to generate the answer as a JSON object containing only the answer character.

As discussed in Section~\ref{sec:intro}, cultural knowledge varies across locations, and we hypothesize that providing geographical context can enhance a model’s reasoning ability in cultural contexts. To test this, we complement our experiments with three different levels of location context \(\ell \in \{\text{none}, \text{region}, \text{region $+$ country}\} \).



\subsection{Zero-shot Experiments} 

Our observations show that evaluation with English prompts outperforms Arabic prompts, consistent with findings from \citet{koto-etal-2024-arabicmmlu, kmainasi2024nativevsnonnativelanguage}. We speculate that this is due to the dominance of English instruction-tuning datasets in LLM development. Therefore, we present the results using English prompts in the main text and include the Arabic results in Appendix~\ref{appendix:arabic_zero_shot}.

\paragraph{Overall Observation}

The overall results presented in Table~\ref{tab:model_scores} reveal notable performance differences between open-weight and closed-weight models in understanding Arabic culture and norms. While some large-scale open-weight models achieve relatively high accuracy—such as Qwen-2.5-Instruct (72B) with 80\%, LLaMA-3.3-Instruct (70B) with 75.4\%, and AceGPT-v2-Chat (32B) with 79.7\%—, closed-wight models represented by GPT-4o demonstrate significantly stronger performance.
Arabic-centric models do not consistently outperform multilingual models. Some, such as Jais, struggle despite being tailored for Arabic, whereas certain multilingual models like Qwen2.5 and Llama-3.3 Instruct surpass them in accuracy.
Within the same model family, performance generally improves with larger model sizes—except for Jais and AceGPT. However, across different model families, scaling does not guarantee better results. This indicates that factors beyond model size, such as pretraining data, architecture, and training recipes, significantly impact cultural comprehension.
When comparing base models to instruction-tuned variants, we observe modest improvements in completion tasks but substantial gains in MCQ tasks. 

While reasoning models have demonstrated impressive capabilities in domains such as mathematics and programming, they perform poorly on our cultural reasoning tasks. This discrepancy suggests that cultural reasoning involves fundamentally different challenges that remain underexplored and require dedicated attention.

Overall, these findings highlight the need for improvements in Arabic and multilingual models to enhance their comprehension of Arabic cultural contexts. Addressing this gap can also help mitigate cultural biases, which have been identified in recent studies, such as the work by \citet{naous-etal-2024-beer}.


\paragraph{Multiple‐Choice (MCQ) Outperforms Completion}

In Table~\ref{tab:model_scores}, we observe that sentence completion is not as reliable as MCQ, despite being a more natural approach that aligns with the sentence completion framework of \datasetname{}. Qwen‑2.5 Instruct (32B), for example, achieves 75.2\% accuracy in MCQ but drops significantly to 37.6\% in sentence completion. Similar disparities are also evident in smaller models; for instance, BLOOMZ (7B) achieves 58.5\% in MCQ but performs at random (31.7\%) in sentence completion. Some models, such as Aya‐Expanse (8B), show only a small gap between MCQ and sentence completion, likely due to pretraining or fine-tuning gaps that hinder their ability to leverage structured prompts effectively. Interestingly, the older version of Llama-3 Instruct (70B) does not benefit from the MCQ strategy, whereas the latest version (Llama-3.3 Instruct) shows a dramatic improvement, increasing from 41\% to 71\%. Given that MCQ yields the best results, we use it for further analysis.


\paragraph{Impact of Location Granularity} Adding finer-grained location information in the prompt does not produce a consistently positive or negative effect on zero-shot performance, yielding mixed results. For instance, when region and country context are included, Jais‐v3 Chat (30B) experiences a 6-point accuracy drop compared to the vanilla prompt ($\ell = \text{None}$). In contrast, Qwen‑2.5 (14B) shows a substantial improvement, increasing from 55.2\% ($\ell = \text{None}$) to 61.6\% when provided with country-level context. These fluctuations suggest that in some models, location specificity does not necessarily enhance cultural understanding.

\begin{table}[t]
\centering
\resizebox{\linewidth}{!}
{\begin{tabular}{lcccccc}
\toprule
\multirow{2}{*}{\textbf{Location}} & \multicolumn{2}{c}{\textbf{GPT-4o}} & \multicolumn{2}{c}{\textbf{Qwen-2.5}} & \multicolumn{2}{c}{\textbf{AceGPT-v2}} \\
\cmidrule(lr){2-3} \cmidrule(lr){4-5} \cmidrule(lr){6-7}
 & \textbf{CS} & $\neg$\textbf{CS} & \textbf{CS} & $\neg$\textbf{CS} & \textbf{CS} & $\neg$\textbf{CS} \\
\midrule

\textbf{Gulf}           & 87.8 & 91.6                         & 72.3 & 80.0                                          & 73.1 & 82.0                                        \\
KSA                     & 88.4 & 91.0                           & \ok 82.1 & 78.9                                        & 77.9 & 81.3                                      \\
UAE                     & \ok 93.0   & 93.4                         & 75.0   & 83.6                                        & 76.0   & 84.2                                      \\
Yemen                   & \no 85.0   & 88.1                         &  66.5 & \no 73.1                                        & 69.4 & 77.6                                      \\ 
\midrule
\textbf{Levant}         & 83.8 & 93.2                         & 69.9 & 86.7                                        & 69.4 & 85.6                                      \\
Lebanon                 & \no 78.8 & \no 82.1                         & \no 61.6 & \no 68.6                                        & \no 63.6 & \no 66.7                                      \\
Syria                   & 85.0   & 93.7                         & 75.0   & 85.4                                        & \no 67.5 & 83.3                                      \\
Palestine               & \ok 95.7 & 94.4                         & 87.0   & 89.2                                        & \ok 87.0   & 90.0                                        \\
Jordan                  & \ok 100.0  & \ok 97.8                         & \ok 90.9 & \ok 95.7                                        & \ok 90.9 & \ok 94.3                                      \\ 
\midrule
\textbf{Nile Valley}    & 87.1 & 97.8                         & 76.8 & 93.3                                        & 74.8 & 88.9                                      \\
Egypt                   & 87.3 & 97.1                         & 78.2 & 91.2                                        & 73.1 & 83.8                                      \\
Sudan                   & 86.8 & \ok 98.2                         & 75.0   & \ok 94.6                                        & 77.1 & \ok 92.0                                        \\ 
\midrule
\textbf{North Africa}   & 86.1 & 86.9                         & 70.8 & 81.1                                        & 73.7 & 79.9                                      \\
Tunisia                 & \no 75.9 & \no 80.9                         &\no  61.4 & \no 69.7                                        & \no 62.7 & \no 72.5                                      \\
Algeria                 & 85.1 & \no 83.8                         & \no 63.5 & 77.2                                        & 68.9 & \no 72.6                                      \\
Morocco                 & 91.3 & \ok 94.2                         & 75.7 & \ok 93.1                                        & 80.6 & \ok 93.6                                      \\
Libya                   & 91.1 & 89.4                         & \ok 81.0   & 85.6                                        & \ok 81.0   & 82.5                                      \\
\bottomrule
\end{tabular}
}\caption{
Performance of \texttt{GPT-4o}, \texttt{Qwen-2.5-72B Instruct}, and \texttt{AceGPT-v2-32B Chat} across countries and regions. \texttt{CS} denotes country-specific examples, while \texttt{$\neg$CS} otherwise. Green cells indicate the top three scores, while red cells highlight the bottom three.
}
\label{tab:english_results_by_country}
\end{table}

\begin{table}[h!]
\centering
\resizebox{\linewidth}{!}{\begin{tabular}{lcccccc}
\toprule
\multirow{2}{*}{\textbf{Topic}} & \multicolumn{2}{c}{\textbf{GPT-4o}} & \multicolumn{2}{c}{\textbf{Qwen-2.5}} & \multicolumn{2}{c}{\textbf{AceGPT-v2}} \\
\cmidrule(lr){2-3} \cmidrule(lr){4-5} \cmidrule(lr){6-7}
 & \textbf{CS} & $\neg$\textbf{CS} & \textbf{CS} & $\neg$\textbf{CS} & \textbf{CS} & $\neg$\textbf{CS} \\
\midrule

Agriculture          & \ok 91.5 & 91.7                         & \ok 83.0   & \ok 90.0                                          & 78.7 & \ok 85.0                                        \\
Art                  & 87.7 & \ok 92.4                         & 74.8 & 84.8                                        & 73.5 & 84.8                                      \\
Daily Act.     & \no 79.4 & \no 90.2                         & \no 65.4 & 83.2                                        & \no 68.4 & \ok 86.6                                      \\
Death                & 84.4 & 91.3                         & 78.1 & \no 83.1                                        & \ok 84.4 & 81.2                                      \\
Family Rel. & \ok 90.0   & 91.1                         & \ok 80.0   & \ok 88.9                                        & 75.0   & 83.3                                      \\
Food                 & 87.0   & 90.4                         & 71.5 & \no 79.4                                        & 70.9 & \no 79.4                                      \\
Habits               & \no 81.8 & 91.0                           & \no 68.8 & 86.2                                        & 71.4 & 82.1                                      \\
Holiday Act.   & 89.2 & \ok 94.4                         & 70.7 & \ok 89.1                                        & \no 66.9 & \ok 90.1                                      \\
Idioms               & 88.9 & 91.9                         & 70.4 & 83.8                                        & \ok 86.4 & 81.1                                      \\
Parenting            & \no 76.2 & \ok 93.8                         & \no 66.7 & 87.5                                        & \no 66.7 & 84.4                                      \\
Trd. Games    & 87.5 & \no 89.6                         & \ok 80.0   & 84.7                                        & 70.0   & \no 80.6                                      \\
Wedding              & \ok 89.4 & \no 89.1                         & 78.0   & \no 79.6                                        & \ok 81.8 & \no 78.9                                      \\
\bottomrule
\end{tabular}
}\caption{
Performance of \texttt{GPT-4o}, \texttt{Qwen-2.5-72B Instruct}, and \texttt{AceGPT-v2-32B Chat} across topics. \texttt{CS} denotes country-specific examples, while \texttt{$\neg$CS} otherwise. Green cells indicate the top three scores, while red cells highlight the bottom three.
}
\label{tab:english_results_by_topic}

\end{table}

\section{Analysis}

\subsection{Result by Categories}
In this section, we expand on our findings based on the top three models from Table~\ref{tab:model_scores}: (1) \texttt{GPT-4o}, (2) \texttt{Qwen-2.5-72B-Instruct}, and (3) \texttt{AceGPT-v2-32B-Chat}. We provide a detailed analysis across country and topic, with each sample categorized as either country-specific (CS) or non-country-specific ($\neg$CS). 

\paragraph{Country} 

In Table~\ref{tab:english_results_by_country}, we observe significant variation in LLM performance across countries, emphasizing the need for country-specific adaptation when deploying models. Questions from Jordan are consistently predicted with high accuracy, exceeding 90\% across all models. However, performance drops significantly for Lebanon and Tunisia, where models struggle to provide correct answers. Even the Arabic-centric AceGPT-v2 achieves only 63.6\% accuracy for Lebanon and 62.7\% for Tunisia. Across the four regions, we find that the Levant is the most challenging, underscoring the difficulty of achieving reliable performance across different cultural and linguistic contexts.

More interestingly, country-specific questions prove to be more challenging than non-country-specific ones across nearly all countries and models. For instance, in the Nile Valley region, GPT-4o experiences an accuracy drop of nearly 10 points when handling country-specific questions, while in the Levant region, Qwen-2.5 sees a 17-point decline. This suggests that when cultural knowledge is not shared across multiple countries or regions, it becomes more distinct and difficult for LLMs to capture accurately.


\begin{figure}[t]
    \centering
    \includegraphics[width=\linewidth]{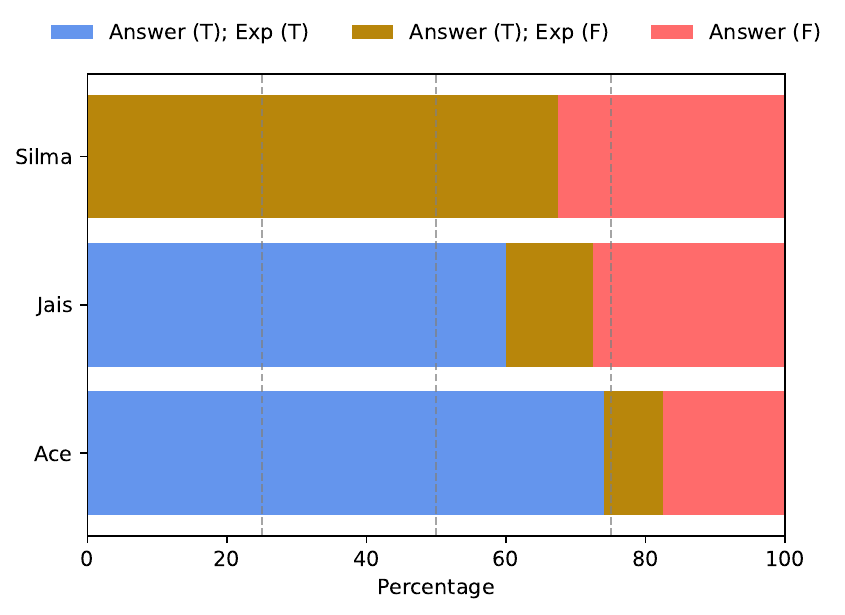}
    \caption{Performance comparison between {\tt jais-30b-chat-v3}, {\tt AceGPT-v2-32B-Chat}, and {\tt SILMA-9B-Instruct-v1.0} based on text generation output. ``Answer (T)'' indicates that the generated
answer is true, while ``Exp (F)'' denotes that the answer explanation is false.}
    \label{fig:generation_exp}
\end{figure}

\paragraph{Topic}
Table~\ref{tab:english_results_by_topic} shows that LLMs encode cultural knowledge differently across various aspects of Arab culture. For example, GPT-4o performs best in agriculture and family relationships, while AceGPT excels in topics related to death and idioms. Meanwhile, Qwen achieves its highest accuracy in agriculture and traditional games. The accuracy gap between the highest- and lowest-performing topics across models ranges from 10 to 20 points, highlighting the difficulty of adapting cultural knowledge in LLMs. Additionally, we observe a consistent trend with Table~\ref{tab:english_results_by_country}, where country-specific samples are more challenging than non-country-specific ones.


\subsection{Can the Model Provide a Reasonable Explanation to Support the Answer?}
\label{sub:explanation_genration}
We focus on Arabic-centric models—Jais, AceGPT, and Silma—to evaluate their actual generation capabilities. For 200 randomly selected samples, we generate responses by appending \<مع ذكر السبب> ("with mentioning the reason") to the Arabic prompt (\S \ref{appendix:zero_shot_prompt}) to instruct the model to provide a brief explanation for its choice. We then manually assess the outputs to verify both the correctness of the answer and the validity of the explanation.

Figure~\ref{fig:generation_exp} presents the results for each model, comparing their generation accuracy with their MCQ performance from Table~\ref{tab:model_scores}. Jais demonstrated a significant improvement, increasing from 40\% in MCQ to 72\% in the manual evaluation. In contrast, Silma’s performance dropped from 73\% in MCQ to 67\% in the generation task. Notably, Silma often failed to generate explanations, instead providing only the answer key or, at times, just the answer text. Meanwhile, AceGPT maintained a consistent performance across both MCQ and generation tests, showing no significant change in accuracy.

\begin{figure}[t]
    \centering
    \includegraphics[width=1\linewidth]{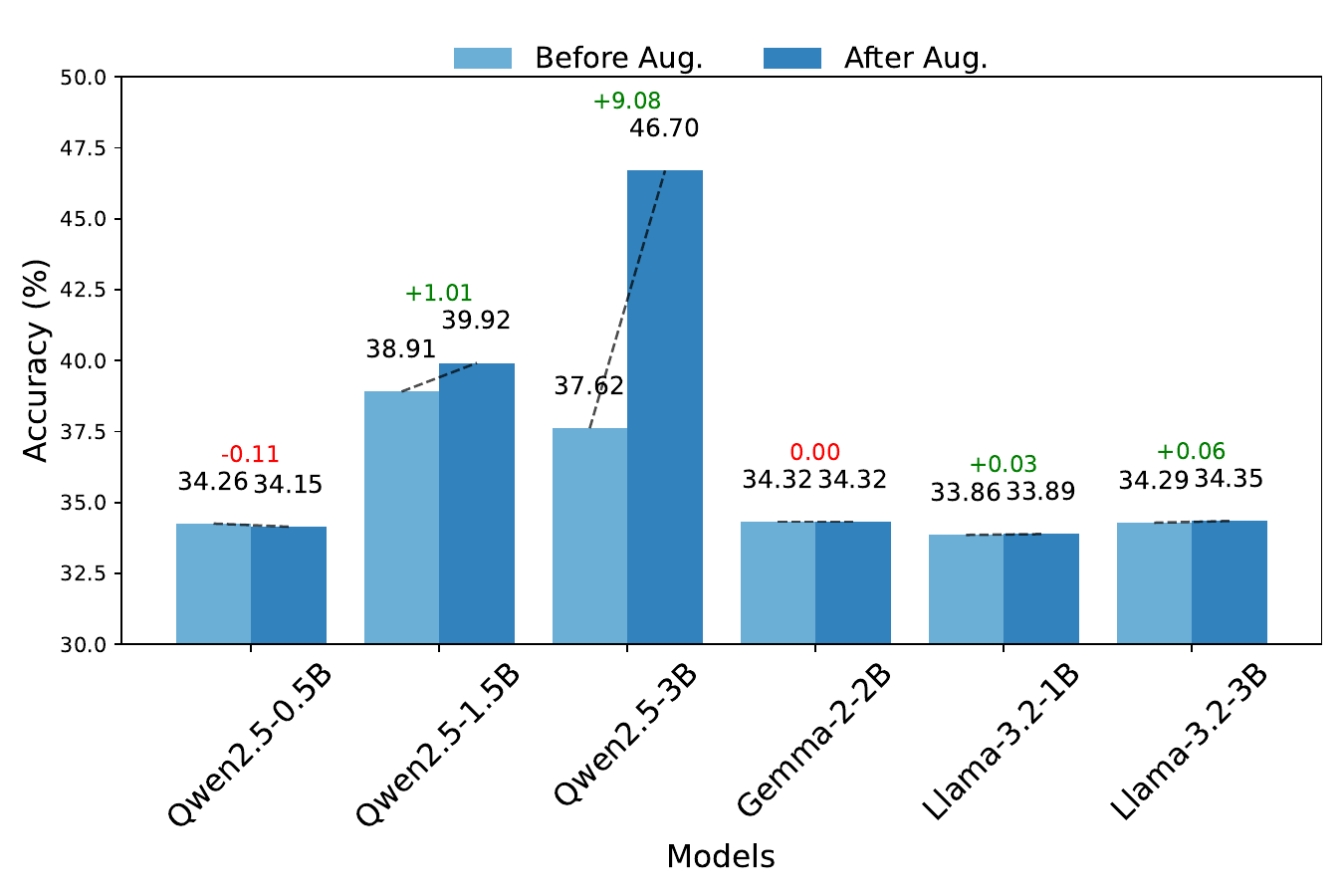}
    \caption{Accuracies per models before and after context augmentation. This experiment uses the Arabic prompt template for MCQ and location \(\ell \in \{\text{region + country}\}\).}
    \label{fig:augment_res}
    \vspace{-0.5cm}
\end{figure}

\subsection{Improving Small Language Model with Additional Context from GPT-4o}

We evaluate six base models with $\leq$3B parameters to assess the impact of cultural context augmentation. Using GPT-4o, we generate five factual Arabic sentences conditioned on the premise, subtopic, and country, following the Arabic prompt in Figure~\ref{fig:prompt_cultural_context}. These sentences are then incorporated into the Arabic multiple-choice question prompt (\S \ref{app:cultural_prompt}) with location context \(\ell \in \{\text{region + country}\} \). Results (Figure \ref{fig:augment_res}) show accuracy gains for Qwen and Llama models, except for Qwen2.5-0.5B (decline) and Gemma-2-2B (unchanged). Qwen2.5-3B achieves the highest improvement across all countries.  These variations likely stem from differences in training data: Qwen and Llama were trained on multilingual datasets, whereas Gemma-2, primarily trained in English, has limited multilingual support.




\section{Conclusion}
We introduced \datasetname{}, a benchmark for evaluating cultural commonsense reasoning in the Arab world. The dataset comprises 3,482 questions across 13 countries, covering 12 daily life domains with 54 fine-grained subtopics, all authored and validated by native speakers. Evaluations on 31 LLMs show significant performance gaps, with open-weight models up to 32B struggling to capture Arab cultural contexts. Variability across countries, regions, and topics highlights the need for more culturally aware models and datasets tailored to the Arabic-speaking world.

\section*{Limitations}

\paragraph{Culture is only one side of reasoning}
While cultural knowledge plays a crucial role in shaping commonsense reasoning, it is only one of several dimensions that contribute to a model's overall reasoning capabilities~\citep{plaat2024reasoninglargelanguagemodels}. Reasoning in LLMs encompasses a broad range of cognitive skills, including logical inference, numerical reasoning, and causal understanding, among others.
\paragraph{The Influence of Dialects on Cultural Reasoning}

The Arab world is characterized by rich dialects that vary not only across countries but also within different regions of the same country. These dialects significantly shape cultural expression, influencing language use in areas such as proverbs, humor, and everyday communication. This is particularly evident in topics like "idioms," where meaning and usage are deeply tied to specific dialects and local linguistic conventions.

However, to ensure that our evaluation isolates cultural commonsense reasoning rather than a model’s proficiency in specific dialects, we constructed \datasetname{} in Modern Standard Arabic (MSA). MSA serves as a unifying linguistic medium across Arabic-speaking countries, allowing us to control for dialectal variation while still capturing essential cultural knowledge. While this approach enhances comparability across regions, it also introduces a limitation: certain cultural concepts that are best expressed through dialect-specific phrasing or context may not be fully represented in our dataset.
\paragraph{Location Leakage} Despite our efforts to systematically control the granularity of location information, some questions or corresponding multiple-choice completions inadvertently reveal the location through the inclusion of location cues (landmarks, national events, etc.) within prompts. Thus, resulting in unintended location leakage, where the model gains access to country-specific cues directly from text rather than controlled contexts, making it difficult to isolate the effect of the location granularity control.

\paragraph{Coverage of All Arab Countries}
While our study covers a significant portion of the Arab world, representing 82\% of the total population, certain unique cultures remain underrepresented. Notably, countries such as Mauritania, Somalia, and Comoros were not included, despite their distinct cultural and linguistic characteristics. These nations, located in North Africa, the Horn of Africa, and the Indian Ocean, respectively, contribute to the broader diversity of the Arab world. Their exclusion was primarily due to the difficulty in sourcing human annotators from these regions.

\section*{Acknowledgments}
We are grateful to the supercomputing center at MBZUAI for providing resources for this research. We also gratefully acknowledge the support of the university for providing funding to hire the human annotators.

\bibliography{acl_latex}

\appendix

\section{Dataset Statement for \datasetname}
\subsection{General Information}
\textbf{Dataset title} \datasetname
\newline
\textbf{Dataset version} 1.0 (Feb 2025)
\subsection{Executive Summary}
\datasetname is a cultural Arabic commonsense reasoning dataset covering 13 Arabic countries in the Gulf, Levant, North Africa, and the Nile Valley. The dataset spans 12 daily life domains and 54 fine-grained subtopics. It was created from scratch by native speakers who validated culturally relevant questions.
\subsection{Curation Rationale}
\datasetname serves as a cultural benchmark to assess large language models' ability to reason within culturally specific contexts. Built from scratch by native speakers, it avoids web-scraped text and undergoes rigorous quality control to ensure lexical accuracy, semantic coherence, and cultural sensitivity.

The dataset creation process involves:
\begin{enumerate}[label=\arabic*.]
    \item \textbf{Coverage Determination:} Selecting relevant countries and topics.
    \item \textbf{Annotator Selection:} Hiring qualified native speakers with deep cultural knowledge.
    \item \textbf{Example Generation:} Each annotator produces 150 examples, with two annotators per country.
    \item \textbf{Cross-Review:} Annotators validate each other’s work by confirming correct answers.
    \item \textbf{Final Review:} Unclear samples are revised or discarded based on comprehensive quality checks.
\end{enumerate}

\subsection{Documentation for Source Datasets}

\datasetname is built entirely from scratch, without relying on web-scraped text, All data is manually created and validated by native speakers from 13 Arabic-speaking countries. 

\subsection{Country and Regional Diversity}
\datasetname covers 13 Arab countries, ensuring rich cultural perspectives.

Our country selection was guided by the goal of broad geographic representation across the Arab world. These 13 countries span four key regions:
\begin{itemize}
    \item \textbf{The Gulf:} Saudi Arabia, Yemen, UAE.
    \item \textbf{The Levant:} Syria, Jordan, Palestine, Lebanon.
    \item \textbf{North Africa:} Morocco, Algeria, Tunisia, Libya.
    \item \textbf{The Nile Valley:} Egypt, Sudan.
\end{itemize}

\subsection{Annotator Demographics}
We recruited 26 expert annotators from 13 Arab countries, with two annotators representing each country. To ensure cultural authenticity and linguistic proficiency, we enforced the following eligibility criteria:
\begin{itemize}
    \item Native Arabic speakers.
    \item Residency in the country for at least 10 years.
    \item Deep understanding of local culture and traditions.
    \item Both parents are native to and reside in the country.
    \item Minimum educational requirement of a high school diploma (higher degrees are preferred).
\end{itemize}
Among the 26 annotators, 14 hold a Bachelor's degree, seven have a Master's degree, two have a PhD, and three have a high school diploma. All annotators participated (attended or watched a record) in an initial online workshop to ensure a clear understanding of the project guidelines.

\subsection{Topic Diversity}
\datasetname features a carefully curated taxonomy of daily life topics. It includes 12 main topics with 54 fine-grained subtopics—covering areas such as food, weddings, holiday activities, daily routines, habits, traditional games, death, art, parenting, agriculture, family relationships, and idioms. This extensive range ensures that the dataset captures both common and unique cultural experiences across the Arab world.

\section{Chosen Topics Distribution}
Table~\ref{tab:topics} shows the distribution of topics and their corresponding subtopics.

\section{Prompts}
\subsection{Zero-shot Experiment prompts}
\label{appendix:zero_shot_prompt}
\begin{figure}[h!]
    \centering
    \includegraphics[width=0.48\textwidth]{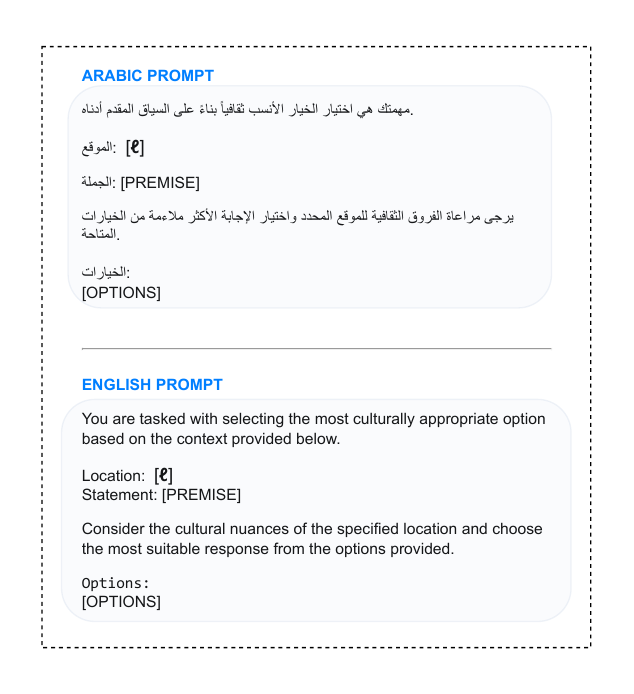}
\caption{Templates for multiple-choice question prompts. Sentence completion prompt template is the same but without the options section. Location \(\ell \in \{\text{none}, \text{region}, \text{region + country}\}\).}
\label{fig:prompts}
\end{figure}

Figure \ref{fig:prompts} shows the prompts in Arabic and English that we used for the zero-shot experiments. The Arabic prompt is also used to generate responses for~\ref{sub:explanation_genration}.

\subsection{Cultural Context prompts}
\label{app:cultural_prompt}
\begin{figure}[h!]
    \centering
    \includegraphics[width=1\linewidth]{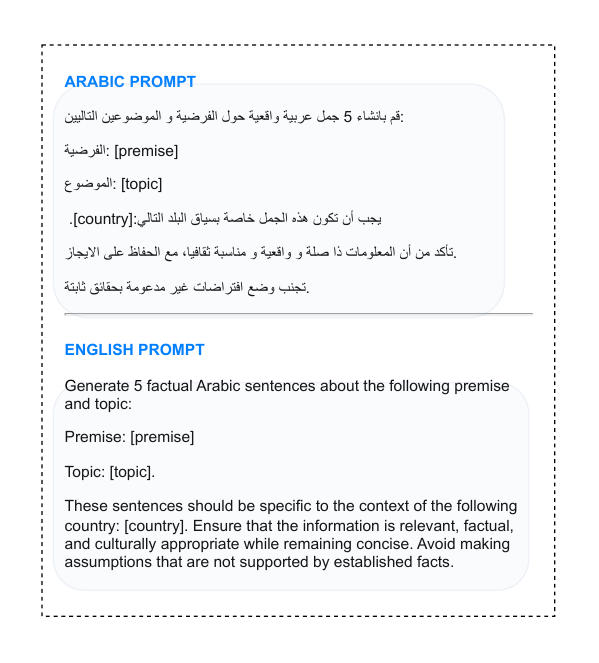}
    \caption{Prompts to generate culturally relevant sentences for context augmentation. The Arabic prompt was used for the generation.}
    \label{fig:prompt_cultural_context}
\end{figure}

Figure~\ref{fig:prompt_cultural_context} displays the Arabic prompt used to generate five culturally grounded sentences for enhancing the small language model with additional context from GPT-4o. An English translation is provided to assist readers unfamiliar with Arabic in understanding the prompt.

\label{appendix:cultural_context_prompt}

\section{Results of the Arabic prompt in the zero-shot experiments}
Table~\ref{tab:model_scores_arabic} presents the zero-shot experiment results using the Arabic prompt, which is illustrated in Figure~\ref{fig:prompts}.
\label{appendix:arabic_zero_shot}

\subsection{Results by Geographic Location - Arabic Prompt}
\label{appendix:arabic_geographic_location_analysis}

\begin{table}[h!]
\centering
\resizebox{\linewidth}{!}{\begin{tabular}{lcccccc}
\toprule
\multirow{2}{*}{\textbf{Topic}} & \multicolumn{2}{c}{\textbf{GPT-4o}} & \multicolumn{2}{c}{\textbf{AceGPT-v2}} & \multicolumn{2}{c}{\textbf{Llama-3.3}} \\
\cmidrule(lr){2-3} \cmidrule(lr){4-5} \cmidrule(lr){6-7}
 & \textbf{CS} & $\neg$\textbf{CS} & \textbf{CS} & $\neg$\textbf{CS} & \textbf{CS} & $\neg$\textbf{CS} \\
\midrule

\textbf{Gulf}           & 81.0   & 92.5                         & 76.3 & 85.3                                     & 69.1 & 81.5                                          \\
KSA                     & 89.5 & 93.4                         & 77.9 & 84.3                                     & 75.8 & 80.1                                          \\
UAE                     & \ok 90.0   & 93.4                         & 75.0   & \ok 90.2                                     & 66.0   & 87.4                                          \\
Yemen                   & \no 72.8 & 88.1                         & 76.2 & 74.6                                     & 67.5 & \no 68.7                                          \\ 
\midrule
\textbf{Levant}         & 77.5 & 92.3                         & \no 67.6 & 86                                       & 66.5 & 84.4                                          \\
Lebanon                 & \no 70.7 & \no 78.8                         & \no 60.6 & \no 67.9                                     & \no 62.6 & \no 66.0                                            \\
Syria                   & 82.5 & 94.6                         & 72.5 & 87.0                                       & 70.0   & 80.8                                          \\
Palestine               & 87.0   & 94.4                         & 78.3 & 87.2                                     & \no 65.2 & 88.8                                          \\
Jordan                  & \ok 100.0  & \ok 96.1                         & \ok 90.9 & \ok 94.3                                     & \ok 90.9 & \ok 93.9                                          \\ 
\midrule
\textbf{Nile Valley}    & 86.8 & \ok 96.1                         & 70.7 & 86.1                                     & 76.0   & 85.0                                            \\
Egypt                   & 88.8 & \ok 98.5                         & 72.1 & 79.4                                     & \ok 76.1 & 72.1                                          \\
Sudan                   & 84.0   & 94.6                         & 68.8 & \ok 90.2                                     & 75.7 & \ok 92.9                                          \\ 
\midrule
\textbf{North Africa}   & 82.0   & 86.0                           & 73.7 & 78.8                                     & 73.5 & 80.9                                          \\
Tunisia                 & \no 67.5 & \no 79.8                         & \no 62.7 & \no 72.5                                     & 68.7 & \no 69.7                                          \\
Algeria                 & 83.8 & \no 84.3                         & 68.9 & \no 73.1                                     & \no 63.5 & 80.2                                          \\
Morocco                 & \ok 90.3 & 94.2                         & \ok 81.6 & \ok 93.1                                     & \ok 83.5 & \ok 93.1                                          \\
Libya                   & 84.8 & 86.2                         & \ok 79.7 & 77.5                                     & 74.7 & 81.2                                          \\
\bottomrule
\end{tabular}
}\caption{
Performance of the best three models using the Arabic prompt. 
The results for {\tt GPT-4o} are using the Region prompt, while the results for {\tt AceGPT-v2-32B-Chat} and {\tt Llama-3.3-70B-Instruct} are using the Country\_Region prompt. 
The first column includes the different locations (countries and regions) with the regions in \textbf{bold}. 
{\tt CS} refers to the Country Specific examples, while {\tt $\neg$CS} refers to the rest of the examples. The green and red cells indicate the top three and bottom three scores, respectively.
}
\label{tab:arabic_results_by_country}
\end{table}

Table~\ref{tab:arabic_results_by_country} presents the country-level breakdown analysis based on the Arabic prompt.
\subsection{Results by Topic - Arabic Prompt}
\label{appendix:arabic_topic_analysis}

\begin{table}[h!]
\centering
\resizebox{\linewidth}{!}{\begin{tabular}{lcccccc}
\toprule
\multirow{2}{*}{\textbf{Topic}} & \multicolumn{2}{c}{\textbf{GPT-4o}} & \multicolumn{2}{c}{\textbf{AceGPT-v2}} & \multicolumn{2}{c}{\textbf{Llama-3.3}} \\
\cmidrule(lr){2-3} \cmidrule(lr){4-5} \cmidrule(lr){6-7}
 & \textbf{CS} & $\neg$\textbf{CS} & \textbf{CS} & $\neg$\textbf{CS} & \textbf{CS} & $\neg$\textbf{CS} \\
\midrule

Agriculture          & \ok 87.2 & \ok 95.0                           & 76.6 & 85.0                                       & \ok 78.7 & 83.3                                          \\
Art                  & \ok 85.2 & 89.9                         & 74.2 & \ok 87.3                                     & 74.8 & \ok 86.1                                          \\
Daily Activities     & \no 76.5 & 89.9                         & \no 67.6 & 83.2                                     & \no 69.9 & 84.3                                          \\
Death                & \no 78.1 & 90.3                         & \ok 87.5 & 84.1                                     & \ok 78.1 & 82.6                                          \\
Family Relationships & 85.0   & 91.1                         & 80.0   & 85.6                                     & \no 60.0   & 85.6                                          \\
Food                 & 82.0   & 91.2                         & 69.3 & \no 79.9                                     & \no 68.7 & \no 76.5                                          \\
Habits               & 80.5 & \no 88.4                         & \no 68.8 & 84.0                                       & 72.7 & 85.8                                          \\
Holiday Activities   & 82.8 & \ok 92.1                         & 74.5 & \ok 88.8                                     & 70.7 & \ok 87.2                                          \\
Idioms               & 84.0   & \ok 97.3                         & \ok 80.2 & \no 78.4                                     & 70.4 & \no 75.7                                          \\
Parenting            & \no 71.4 & 91.7                         & 71.4 & 84.4                                     & \ok 76.2 & \ok 88.5                                          \\
Traditional Games    & \ok 86.2 & \no 89.6                         & \no 65.0   & \no 77.8                                     & 72.5 & \no 77.8                                          \\
Wedding              & 84.1 & \no 89.8                         & \ok 80.3 & \ok 85.7                                     & 75.8 & 81.6                                          \\
\bottomrule
\end{tabular}
}\caption{
Performance of the best three models using the Arabic prompt across the different Topics.
The results for {\tt GPT-4o} are using the Region prompt, while the results for {\tt AceGPT-v2-32B-Chat} and {\tt Llama-3.3-70B-Instruct} are using the Country\_Region prompt.
{\tt CS} refers to the Country Specific examples, while {\tt $\neg$CS} refers to the rest of the examples. The green and red cells indicate the top three and bottom three scores, respectively.}
\label{tab:arabic_results_by_topic}
\end{table}

Table~\ref{tab:arabic_results_by_topic} presents the breakdown analysis by topic based on the Arabic prompt.
\begin{table*}[ht]
\centering
\resizebox{0.8\textwidth}{!}{\begin{tabular}{p{4cm} p{9cm} c}
\hline
\textbf{Topics}           & \textbf{Sub-topics}                                                                                                                                                                                                                     & \textbf{\#Samples} \\ \hline
Food                     & Breakfast (5), Lunch (5), Dinner (2), Sahoor (Ramadan) (5), Iftar (Ramadan) (5), Dessert (3), Fruits (3), Snacks (2)                                                                             & 30                 \\ \hline
Wedding                  & Wedding location (1), Wedding food (1), Wedding dowry (1), Wedding other logistics (2), Men ceremony vs. women ceremony (2), Songs and activities during the wedding (5)                                                               & 12                 \\ \hline
Holiday Activities       & Traditions before religious holidays (5), Traditions during religious holidays (10), Activities for non-religious holidays (5)                                                                                                           & 20                 \\ \hline
Daily Activities         & Before going to work/college (4), While on the way to college/work (3), Things you do with colleagues/friends while at work/college (2), Things you do after coming back from work/uni (men) (2), Things you do after coming back from work/uni (women) (2), Household activities (groceries, fixing things, cleaning, etc.) (6), Things you do in your free time (indoors or outdoors) (5) & 24                 \\ \hline
Habits                   & Eating habits (3), Stereotypes (5), Communication habits (3), Financial habits (1), Gift-giving practices (1), Cleanliness habits (2)                                                                                                     & 14                 \\ \hline
Traditional Games        & Childhood games indoors (5), Childhood games outdoors (5)                                                                                                                                                                                & 10                 \\ \hline
Death                    & Before burying (3), Burying rituals (2), After burying ceremonies (4), Inheritance (1)                                                                                                                                                   & 10                 \\ \hline
Art                      & Musical instruments (3), Local songs (3), Local dances (4)                                                                                                                                                                               & 10                 \\ \hline
Parenting                & Parents-child actions (3), Grandparents-child actions (2)                                                                                                                                                                                 & 5                  \\ \hline
Agriculture              & What to plant (3), While planting (1), Harvest (1)                                                                                                                                                                                        & 5                  \\ \hline
Idioms                   & Idioms in context (5)                                                                                                                                                                                                                                & 5                  \\ \hline
Family Relationships     & Between siblings (2), With cousins (1), Relationship of parents and child (2)                                                                                                                                                            & 5                  \\ \hline
\end{tabular}
}\caption{Overview of topics, sub-topics, and the sample counts for each topic. The number in parenthesis beside each subtopic represents the number of samples for each subtopic.}
\label{tab:topics}
\end{table*}

\begin{table*}[h!]
\centering
\resizebox{0.8\textwidth}{!}{%
\begin{tabular}{lcccccc}
\hline
\multirow{2}{*}{\textbf{Model (\#parameter)}} &

  \multicolumn{3}{c}{\cellcolor{blue!7}\textbf{Completion}} &
  \multicolumn{3}{c}{\cellcolor{red!7}\textbf{MCQ}} \\ \cline{2-7} 
                             & \cellcolor{blue!7}$\ell =$ None    & \cellcolor{blue!7}$\ell =$ R  & \cellcolor{blue!7}$\ell =$ R + C & \cellcolor{red!7}$\ell =$ None    & \cellcolor{red!7}$\ell =$ R  & \cellcolor{red!7}$\ell =$ R + C \\ \hline
Human                        & $-$           & $-$           & 100.0         & $-$           & $-$           & 100.0         \\
Random                                        & 33.3          & 33.3            & 33.3                      & 33.3          & 33.3            & 33.3                       \\ 
\hline
BLOOMZ (7B)                                   & 30.1                 & 30.7                & 30.9                     & 50.6                & 52.2                & 52.7                      \\ 
\hdashline
mT0$_xxl$ (14B)                               & 26.6                 & 26.3                 & 26.9                      & 65.5                 & 66.3                 & 66.4                       \\ 
\hdashline
Llama-3.1 (8B)                                & 27.7                 & 27.5                 & 27.6                      & 34.3                 & 34.1                 & 34.2                       \\
Llama-3.1 Instruct (8B)                       & 32.2                 & 31.0                   & 31.3                      & 37.8                 & 36.8                 & 37.8                       \\
Llama-3 Instruct (70B)                        & 36.6                 & 37.5                 & 38.5                      & 47.1                 & 37.0                   & 38.7                       \\
Llama-3.3 Instruct (70B)                      & 39.0                   & 39.6                 & 39.9                      & \textbf{78.4}        & \textbf{77.8}        & \textbf{78.8}              \\ 
\hdashline
Aya-Expanse (8B)                              & 34.7                & 35.8                & 36.9                     & 36.3                & 38.1                & 39.3                      \\
Aya-Expanse (32B)                             & 37.9                 & 39.3                 & 39.9                      & 38.4                 & 43.7                 & 44.6                       \\ 
\hdashline
Gemma-2 (9B)                                  & 32.0                & 32.2                 & 32.7                     & 34.5                & 35.8                & 35.4                      \\
Gemma-2 Instruct (9B)                         & 32.5                & 33.5                & 33.5                     & 34.4                & 34.3                & 34.3                      \\
Gemma-2 (27B)                                 & 33.8                 & 34.4                & 35.0                     & 34.3                & 34.3                & 34.4                      \\
Gemma-2 Instruct (27B)                        & 35.9                & 36.4                & 37.1                     & 34.4                & 34.9                & 34.9                      \\ 
\hdashline
Qwen2.5 (7B)                                  & 30.0                   & 30.4                & 30.4                     & 47.9                & 47.0                & 47.7                      \\
Qwen2.5 Instruct (7B)                         & 32.5                & 33.2                & 33.8                     & 51.6                & 37.7                & 39.3                      \\
Qwen2.5 (14B)                                 & 32.4                 & 33.1                 & 33.5                      & 46.5                 & 57.8                 & 57.4                       \\
Qwen2.5 Instruct (14B)                        & 36.5                 & 37.7                 & 37.2                      & 52.2                 & 58.4                 & 59.5                       \\
Qwen2.5 (32B)                                 & 33.4                & 34.2                 & 34.5                     & 42.0                & 43.7                & 42.7                      \\
Qwen2.5 Instruct (32B)                        & 36.6                & 37.7                & 37.8                     & 70.7                & 74.6                & 76.3                      \\
Qwen2.5 (72B)                                 & 35.4                & 36.2                & 36.4                     & 48.0                & 52.2                & 57.3                      \\
Qwen2.5 Instruct (72B)                        & \textbf{39.6}       & \textbf{40.1}       & \textbf{41.0}            & 61.5                & 64.5                 & 65.4                      \\ 

\hline
DeepSeek-R1-Distill-Llama (70B) & 36.7 & 37.1 & 37.6 & 34.7 & 34.5 & 35.0 \\
DeepSeek-R1-Distill-Qwen (32B)  & 33.6 & 34.4 & 35.0 & 34.3 & 34.3 & 34.3 \\
QwQ (32B)                        & 22.83 & 22.29 & 22.77 & 32.88 & 32.62 & 32.28 \\

\hline
Jais (13B)                                    & 37.9                 & 37.8                 & 38.3                      & 33.9                 & 33.6                 & 33.8                       \\
Jais chat (13B)                               & 39.8                 & 39.9                 & 40.6                      & 40.7                 & 39.6                 & 38.4                       \\
Jais-v3 (30B)                                 & 39.9                 & 40.4                 & 40.6                      & 34.8                 & 35.8                 & 35.6                       \\
Jais-v3 Chat (30B)                            & 34.0                   & 34.0                   & 34.5                      & 35.2                 & 36.2                 & 39.7                       \\ 
\hdashline
SILMA Instruct (9B)                           & 32.5                 & 33.3                 & 33.4                      & 70.2                 & 70.7                 & 70.7                       \\ 
\hdashline
AceGPT-v2 (8B)                                & 29.9                & 31.2                & 31.6                     & 34.3                & 34.2                & 34.2                      \\
AceGPT-v2 Chat (8B)                           & 34.6                & 35.1                & 35.9                     & 44.8                 & 45.0                   & 45.4                       \\
AceGPT-v2 Chat (32B)                          & 37.7                & 38.9                & 39.0                        & \textbf{78.8}       & \textbf{78.2}       & \textbf{79.8}             \\
AceGPT-v2 Chat (70B)                          & \textbf{42.9}       & \textbf{44.5}       & \textbf{45.1}            & 73.6                & 73.0                & 74.4
\\ 
\hdashline
ALLaM-Instruct-preview (7B)                           & 36.5                 &37.2                 & 37.9                      & 70.0                 & 71.9                 & 74.4                       \\ 
\hline
GPT-4o                                        &                      &                      &                           & \textbf{88.5}        & \textbf{91.9}        & \textbf{91.1}              \\ 
\hline

\hline
\end{tabular}%
}
\caption{Zero-shot accuracy results for the Arabic prompt across various models and settings. ''MCQ`` refers to the multiple-choice question evaluation method, and  $\ell$ represents the inclusion of location context (''R`` indicates the region, and ''C`` denotes the corresponding country). Bolded numbers highlight the highest score within each model group.}
\label{tab:model_scores_arabic}
\end{table*}
\end{document}